\documentclass[12pt]{article}
\usepackage[utf8]{inputenc}
\usepackage[T1]{fontenc}
\usepackage[english]{babel}
\usepackage{ifpdf,newtxtext,newtxmath} 
\usepackage{array,graphicx,dcolumn,multirow,hevea,abstract,hanging,fancyhdr,float}
\newcommand{\jref}{error}
\newcommand{\jhead}{Open publication}
\newcommand{\jdate}{January 2021}
\fancypagestyle{firstpage}{%
 \lhead{\protect\small \jhead}
 \rhead{}
}
\usepackage[labelfont=sc,textfont=sf]{caption}
\usepackage[hyperfootnotes=false,breaklinks=true]{hyperref} 
\usepackage[hyphenbreaks]{breakurl}

\usepackage{booktabs} 
\usepackage{siunitx}
\usepackage{gensymb}
\newcolumntype{L}[1]{>{\raggedright\arraybackslash }p{#1}} 
\newcolumntype{C}[1]{>{\centering\arraybackslash }p{#1}}
\newcolumntype{R}[1]{>{\raggedleft\arraybackslash }p{#1}}
\newcolumntype{d}[1]{D{.}{.}{#1}} 

\let\tempone\itemize
\let\temptwo\enditemize
\let\tempthree\enumerate
\let\tempfour\endenumerate
\renewenvironment{itemize}{\tempone\setlength{\itemsep}{0pt}}{\temptwo}
\renewenvironment{enumerate}{\tempthree\setlength{\itemsep}{0pt}}{\tempfour}

\newcommand\episodelength{$10$}
\newcommand\batchsize{$200$}
\newcommand\panrangemin{-10}
\newcommand\panrangemax{10}
\newcommand\tiltrangemin{-10}
\newcommand\tiltrangemax{10}
\newcommand\panrange{$\left[\panrangemin\degree, \panrangemax\degree\right]$}
\newcommand\tiltrange{$\left[\tiltrangemin\degree, \tiltrangemax\degree\right]$}
\newcommand\vergencerangemin{0}
\newcommand\vergencerangemax{20}
\newcommand\vergencerange{$\left[\vergencerangemin\degree, \vergencerangemax\degree\right]$}
\newcommand\discountfactorval{0}
\newcommand\discountfactor{$\discountfactorval$}
\newcommand\epsilongreedy{$0.05$}
\newcommand\clr{$5.10^{-4}$}
\newcommand\mlr{$5.10^{-4}$}
\newcommand\buffersize{$1000$}
\newcommand\minscreenspeed{$0$}
\newcommand\maxscreenspeed{$4$}
\newcommand\minscreendistance{$0.5$}
\newcommand\maxscreendistance{$5$}
\newcommand\interocculardistance{$6.8$}
\newcommand\rewardscaling{$600$}
\newcommand\pxangle{\si{px}}
\newcommand\pxangularspeedunit{\si{px/iteration}}
\newcommand\pxangularaccelerationunit{\si{px/iteration^2}}

\title{Self-Calibrating Active Binocular Vision via Active\\
Efficient Coding with Deep Autoencoders}

\author{
Charles Wilmot\thanks{\textit{Frankfurt Institute for Advanced Studies}
Frankfurt, Germany Email: wilmot@fias.uni-frankfurt.de}
\and 
  Bertram E.~Shi\thanks{\textit{Hong Kong University of Science and Technology} Clear Water Bay, Hong Kong Email: eebert@ust.hk}
\and
  Jochen Triesch\thanks{\textit{Frankfurt Institute for Advanced Studies} Frankfurt, Germany Email: triesch@fias.uni-frankfurt.de}
}

\date{} 
\begin{document} 

\begin{htmlonly}
\href{\jref}{\jhead}, \jdate, pp.\
\end{htmlonly}

\maketitle
\thispagestyle{firstpage}

\begin{abstract}

We present a model of the self-calibration of active binocular vision comprising the simultaneous learning of visual representations, vergence, and pursuit eye movements. The model follows the principle of Active Efficient Coding (AEC), a recent extension of the classic Efficient Coding Hypothesis to active perception. In contrast to previous AEC models, the present model uses deep autoencoders to learn sensory representations. We also propose a new formulation of the intrinsic motivation signal that guides the learning of behavior. We demonstrate the performance of the model in simulations.

\end{abstract}


\setlength{\baselineskip}{16pt plus.2pt}

\section{Introduction}


Human vision is an active process and comprises a number of different types of eye movements. How the human visual system calibrates itself and learns the required sensory representations of the visual input signals is only poorly understood. A better understanding of this process might allow to build fully self-calibrating active vision systems that are robust to perturbations, e.g., \cite{zhao2013}, which could in turn form the basis for models describing the autonomous learning of object manipulation skills such as reaching or grasping, e.g., \cite{bourdonnaye2018}. Here, we present a model of the self-calibration of active binocular vision that formulates the task as an intrinsically motivated reinforcement learning problem. In contrast to classic computer vision solutions to vergence control and object tracking, our model does not require pre-defined visual representations or kinematic models. Instead, it learns ``from scratch'' from raw images. Nevertheless, it achieves sub-pixel accuracy in its vergence and tracking movements demonstrating successful self-calibration.


\subsection{Biological Background} Humans and many other species have two forward facing eyes providing two largely overlapping views of the world. Initially, the visual signals are transformed to electrical signals in the retina. Different types of retinal ganglion cells are responsible for transmitting different kinds of information to the brain. In particular, the so-called magnocellular pathway conveys information at high temporal resolution required for motion vision \cite{jeffries}, while the so-called parvo-cellular pathway has less temporal resolution but color sensitivity and higher spatial resolution \cite{pavro}. From the retina, information is passed to the lateral geniculate nucleus, where information from both eyes is kept separate. Only in the primary visual cortex, the next processing stage, individual neurons receive information from both eyes. In particular, there are cells that detect small differences in local image structures at corresponding retinal locations in the left and right eye, so-called binocular disparities \cite{binocular}. Furthermore, there are neurons that detect local image motion \cite{motion}. How the response properties of primary visual cortex cells develop has been the subject of a large body of research\cite{simplecells1, binocular, simplecells2}. A widely accepted view is that these representations reflect an optimization of the visual system towards coding efficiency.

\subsection{Efficient Coding in Perception} In particular, inspired by information theory, Horace Barlow proposed a model of sensory coding postulating that neurons minimize the number of spikes needed for transmitting sensory information \cite{barlow1961}. This would help to save energy, which is highly relevant, since the brain has high metabolic demands. It has been argued in \cite{zhaoping2007}, that retinal receptors can receive information at a rate of $10^9 {\rm bit}/s$ \cite{kelly1962}, while the optic nerve can only transmit information at $10^7 {\rm bit}/s$ \cite{nirenberg2001}. This implies that the sensory information must be substantially compressed. Based on the idea of finding efficient codes for sensory signals, a large number of models have been proposed to explain the shapes of receptive fields in sensory brain areas for different modalities (vision, audition, olfaction, touch). More recently, it has been argued that adaptation of the organism's {\em behavior} can also help to make the coding of sensory information more efficient. This theory is called Active Efficient Coding (AEC) \cite{zhao2013, teuliere2014, eckmann2019}. It models the self-calibration of sensorimotor loops, where sensory input statistics shape the sensory representation, the sensory representation shapes the behavior and the behavior in turn shapes the input statistics. AEC has mostly been studied in the context of vision. There, it has been shown that AEC models can account for the self-calibration of active stereo vision, active motion-vision, and accommodation control or combinations thereof. These models have used only shallow neural network architectures to learn to encode the sensory signals. Here, we investigate potential benefits of deeper network architectures by utilizing deep autoencoders and formulate a new intrinsic reward signal to simultaneously learn the control of vergence and pursuit eye movements through reinforcement learning. Our results show that the model achieves sub-pixel accuracy in a simulated agent in a 3-D environment.

\section{The Model}

\begin{figure}
\includegraphics[width=\textwidth]{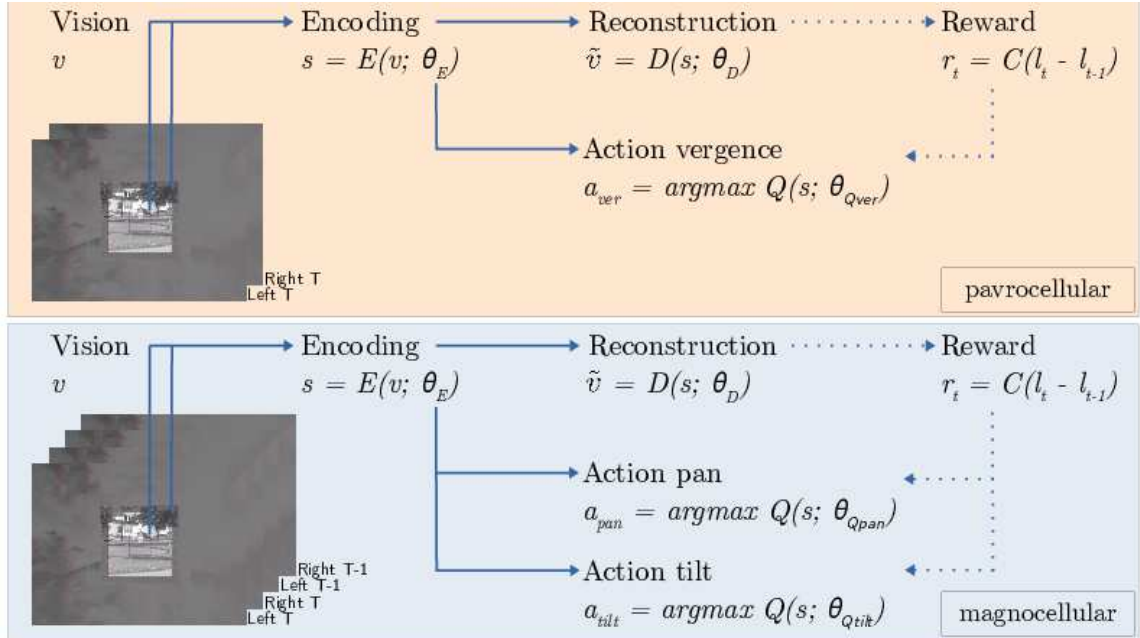} 
\caption{Architecture of the model. Two regions are extracted at the center of the left / right camera images and encoded. The condensed representation is used to train the $Q$-function. The latter is trained to maximize the reward, which is proportional to the improvement of the reconstruction error of the encoder.}
\label{schema}
\end{figure}

\subsection{Sensory encoding via deep autoencoders}

When the two eyes verge on the same point, the foveal regions of the two retinal images become more and more similar. As a consequence, the mutual information between the left and right foveal image representations  ${\rm MI}(I_{\rm L}; I_{\rm R})$ for a given disparity $d$ increases as $d$ goes to $0$. It indicates how redundant the left and right images are and reflects the quality of the fixation. Similarly, tracking a moving object can be achieved by maximizing the information redundancy of the foveal image region across time by maximizing ${\rm MI}(I_t ;I_{t-1})$ (for one or more eyes).


Here, we propose to measure the redundancy in the visual data via training an auto-encoder, hypothesizing that an information stream is better reconstructed when it is more redundant (see measurements in Figs.~\ref{vshaped1} and \ref{vshaped2}). The first advantage of this technique is that it is agnostic to the underlying data representation (for example the RGB channels in left and right data streams could be expressed in different bases or be non-linearly transformed, as only the data redundancy truly matters). Potentially, the algorithm could also exploit highly non-linear redundancies between different sensory modalities, given that the encoding network is sufficiently deep such that it captures these redundancies. The second advantage of this technique is that it learns a condensed representation of the sensory information which can be used by other learning components of the system. Such lossy compression of information may be essential for learning abstract representations at higher processing levels.



We consider a binocular vision system with $3$ degrees of freedom: the pan and the tilt control conjugate horizontal and vertical movements of the gaze and the vergence controls inward and outward movement of the eyes with opposite sign across the two eyes. To properly fixate objects, the agent controlling the vergence must increase the redundancy in the left and right camera images, whereas the agents controlling the pan and tilt must increase the redundancy between consecutive images.
We therefore used different inputs for the vergence, and for the pan and tilt agents. The visual sensory information for the vergence agent consists of the left and right images concatenated on the color dimension, whereas that of the pan and tilt joints consists of the concatenation of left and right images at time $t$ and $t - 1$. The processing taking place on these inputs is the same for all agents. The different visual inputs for vergence vs.\ pan and tilt control reflect the distinction of two separate visual pathways discussed above: the magnocellular pathway with greater temporal resolution and lower spatial resolution compared to the parvocellular pathway.

Let $v(t)$ denote one of the two visual sensory information streams (in the following we will drop the explicit time dependence for compactness of notation), and $E$ and $D$ be, respectively, the encoder and decoder parts of an auto-encoder, such that
\begin{align}
    s &= E\left(v; \theta_E\right) \; \text{and}\\
    \tilde{v} &= D\left(s; \theta_D\right) \; ,
\end{align}
with $s$ representing the encoding and $\tilde{v}$ its reconstruction. We use the loss function $l = \frac{1}{3 N_\text{pixels}}\sum_{ijk}(v_{ijk} - \tilde{v}_{ijk})^2$ for training the encoder and decoder weights $\theta_E$ and $\theta_D$, where $i$, $j$ and $k$ index the height, width, and color dimensions.


\label{v_shape_text}
Both eyes see the world from a slightly shifted perspective. The apparent shift of an object on the retinal image is called binocular disparity. In this paper we measure it in pixels.
In order for the vergence control to work, the encoding / decoding component needs to learn a representation such that increasing binocular disparities induce a degradation of the reconstruction quality. While this is the case for all network architectures we tried, it is important to verify that the range of binocular disparities at which this is true matches the range of disparities at which we want the system to operate. For example, if we want the system to have a fixation accuracy better than $1$ pixel, we must check that the reconstruction error at $1$ pixel disparity is greater than the reconstruction error at $0$ pixel disparity. Similarly, if we want the agent to be capable of resolving disparities greater than $10$ pixels, we want the corresponding reconstruction error to be greater than for lower disparities. Since we want the model to operate over a wide range of disparities (and object velocities) we encode the visual input at different spatial scales, much like the retina samples the world at lower resolution towards the periphery. Details are given in Section~\ref{experimental-setup}.

\subsection{Learning of the Behavior Component}
\subsubsection{Intrinsically motivated reinforcement learning formulation}
We consider the classical Markov decision process framework, where at discrete time $t = 0, 1, 2, \ldots $ an agent observes sensory information $s_t = E\left(v_t, \theta_E\right)$ and chooses action $a_t$ according to the distribution $\pi\left(a_t | s_t\right)$. After applying the action, the agent transitions to a new state according to a transition function $s_{t+1} = T(s_t, a_t)$, and receives a reward $r_t$. While reinforcement learning classically considers a reward provided by the agent's environment through a potentially stochastic reward model, we here define an intrinsic reward based on the agent's sensory encoding of its environment. Specifically, we define the reward

\begin{equation}
    \label{new-reward}
    r^{\rm new}_t = C\left(l_t - l_{t+1}\right) \; ,
\end{equation}
where $C$ is a scaling factor. This reward signal measures the {\em improvement} of encoding quality, i.e., it favors movements that cause transitions from high to low reconstruction error of the autoencoder representing the visual input. We also compare the training speed obtained when training the agent with the simpler reward
\begin{equation}
    \label{old-reward}
    r^{\rm old}_t = - C l_{t+1} \; ,
\end{equation}
which has been used in previous AEC models and simply measures the quality of the encoding.

The goal of reinforcement learning is to learn a policy function $\pi$ that maximizes the (discounted) sum of future rewards $R_t$ called return

\begin{equation}
    R_t = \sum_{i=0}^\infty \gamma^i r_{t + i}
\end{equation}

Where $\gamma$ is a discount factor in $\left[0, 1\right]$ ensuring the convergence of the reward sum. In this particular application of reinforcement learning, the agent does not need to plan ahead his behaviour, consistent with our observation that the algorithm works best for $\gamma = \discountfactorval$.

\subsubsection{RL-Algorithm} We opted for an asynchronous version of the DQN algorithm \cite{mnih2013, mnih2016}. It consists of a $Q$-value function approximation

\begin{equation}
    q = \begin{pmatrix}
           q_0 \\
           \vdots \\
           q_n
         \end{pmatrix} = Q\left(s, \theta_Q\right) \; ,
\end{equation}

where $q_j$ represents an estimate of the return after performing discrete action $j$ in state $s$ and $n$ is the number of possible actions. The loss for training the $Q$-function is the Huber loss between the estimate and the return target \cite{huber1964}. Exploration during the training phase is performed via an $\epsilon$-greedy sampling.

\begin{table}\centering
    \caption{Networks architecture}
    \begin{tabular}{rl}\toprule
        Network & Architecture \\
        \midrule
        Encoder    & conv $96$ filters, size $8 \times 8$ stride $4$\\
        (vergence) & conv $24$ filters, size $1 \times 1$ stride $1$\\
                   & \\
        Decoder    & conv $384$ filters, size $1 \times 1$ stride $1$\\
        (vergence) & \\
                   & \\
        Encoder    & conv $192$ filters, size $8 \times 8$ stride $4$\\
        (pan, tilt)& conv $48$ filters, size $1 \times 1$ stride $1$\\
                   & \\
        Decoder    & conv $768$ filters, size $1 \times 1$ stride $1$\\
        (pan, tilt)& \\
                   & \\
        Critic     & conv $2 \times 2$ stride $1$\\
                   & max-pooling $2 \times 2$ stride $2$\\
                   & flatten\\
                   & concatenate all scales\\
                   & fully-connected $200$\\
                   & fully-connected $9$\\
        \bottomrule
    \end{tabular}
    \label{tab:net_arch}
\end{table}

\begin{table}
    \caption{Parameters value}
    \centering
    \begin{tabular}{rl}\toprule
        Parameter & Value \\
        \midrule
        Pan range & \panrange{}\\
        Tilt range & \tiltrange{}\\
        Vergence range & \vergencerange{}\\
        Discount factor $\gamma$ & \discountfactor{}\\
        Encoder learning rate & \mlr{}\\
        Critic learning rate & \clr{}\\
        Episode length & \episodelength{} iterations\\
        Buffer size & \buffersize{} transitions\\
        Batch size & \batchsize{} transitions\\
        Epsilon $\epsilon$ & \epsilongreedy{}\\
        Reward scaling factor $C$ & \rewardscaling{}\\
        \bottomrule
    \end{tabular}
    \label{tab:parameters}
\end{table}

\section{Experimental setup}
\label{experimental-setup}

We conducted our experiments using the robot simulator CoppeliaSim (previously named V-REP) using the python API PyRep \cite{james2019}, within which a robot head composed of two cameras separated by \interocculardistance{} cm was simulated. Each camera has a resolution of $240 \times 320$ \pxangle{} (height $\times$ width) for a horizontal field of view of $90\degree$ (therefore $1$ pixel $= 0.28 \degree$). To make the results easier to interpret, we expressed all angles and angular velocities in \pxangle{} and \pxangularspeedunit{}, respectively.
In the simulated environment, a screen was moving at uniform speeds varying from \minscreenspeed{} to \maxscreenspeed{} \pxangularspeedunit{} in front of the robot head, at distances between \minscreendistance{} and \maxscreendistance{} meters. 
The screen displayed natural stimuli taken from the \textit{McGill Calibrated Color Image Database} \cite{olmos2004}.

The two spatial scales of visual processing of the cameras are realized by extracting two centered $32 \times 32$ pixel regions per camera (cf.\ \figurename~\ref{schema}), respectively covering a field of view of $9\degree$ (fine scale) and $27\degree$ (coarse scale).
The auto-encoder for each scale corresponds to a $3$-layered fully-connected network encoding patches of size $8 \times 8$ pixels. This patch-wise autoencoder is implemented as a convolutional neural network with filter size $8 \times 8$ in the first layer and $1 \times 1$ in the following layers (see Tab.~\ref{tab:net_arch}).
Figure~\ref{vshaped2} shows the reconstruction error as a function of the binocular disparity for each scale after learning.

The critic network $Q$ can be described in $2$ parts. The first part operates individually on each scale. It is composed of a convolutional layer followed by a pooling layer. The results are then flattened and concatenated before being processed by $2$ fully-connected layers in the second part (cf.\ Tab.~\ref{tab:net_arch}).

All networks are trained using the Adam algorithm \cite{adam} with a learning rate of \mlr{}. We use a value of $\epsilon = \; \epsilongreedy{}$ for the $\epsilon$-greedy sampling.
The training is divided into episodes of \episodelength{} iterations. Each time an episode is simulated, all its transitions are placed in a replay buffer of size 1000 and a batch of data is then sampled uniformly at random from the buffer for training the networks. We use a batch size of \batchsize{}.
The training is spread over multiple processes, each simulating one agent. Each process asynchronously pulls the current weight values from a server, uses them to compute the weight updates $\Delta\theta$, and sends them to the server, which is responsible for performing the updates.

The robot has $3$ joints available to control the eyes. All joints used the same action discretization. However, the vergence joint operates in velocity control mode, while the pan and tilt actions are interpreted as accelerations. The action set we used for all joints is the following: $\left[-4, -2, -1, -\frac{1}{2}, 0, \frac{1}{2}, 1, 2, 4\right]$ \pxangularspeedunit{} (vergence), or \pxangularaccelerationunit{} (pan and tilt). The vergence angle of the robot's eyes is constrained between $\vergencerangemin \degree$ (parallel optical axes of the two eyes) and $\vergencerangemax \degree$ (inward rotated eyes). The pan and tilt joints are constrained to remain in \mbox{\panrange{}.}


At regular intervals, the training is paused, and the agents' performances are measured.
For evaluating the agents' performances, we gather two sets of data at each testing step. One, the \textit{controlled-error set}, gauges the performance of the agents under defined apparent disparities. The other, the \textit{behaviour set}, measures how the policies of the agents recover from initial disparities. All measurements are repeated for $20$ stimuli displayed on the screen $2$ \si{m} away from the eyes of the robot. 
To construct the controlled-error set, we simulated various pan, tilt and vergence errors by manually setting the speed of the screen and the vergence angle of the eyes. Only one joint was tested at a time, meaning that the errors for the two others were set to $0$. We then recorded the reconstruction errors of the fine and coarse scales and the agents' preferred actions. 
The behaviour set is the recording of $20$ iterations of the agents' behaviour, starting from controlled initial pan, tilt, or vergence errors.

\section{Results}

\begin{figure}
\centerline{\includegraphics[width=0.95 \linewidth]{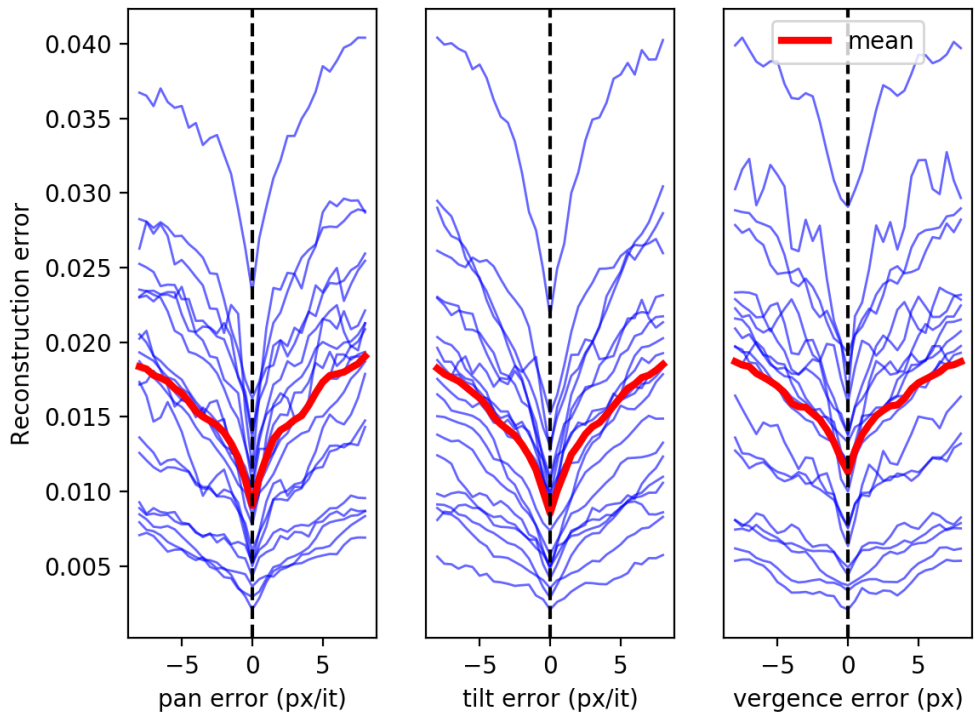}}
\caption{The autoencoders' reconstruction errors averaged across the two scales as a function of the pan, tilt, and vergence error. Each blue curve corresponds to a different stimulus displayed on the screen. The red curve represents the mean. For each plot, the error for the two other joints is set to $0$.}
\label{vshaped1}
\end{figure}

\begin{figure}
\centerline{\includegraphics[width=0.95 \linewidth]{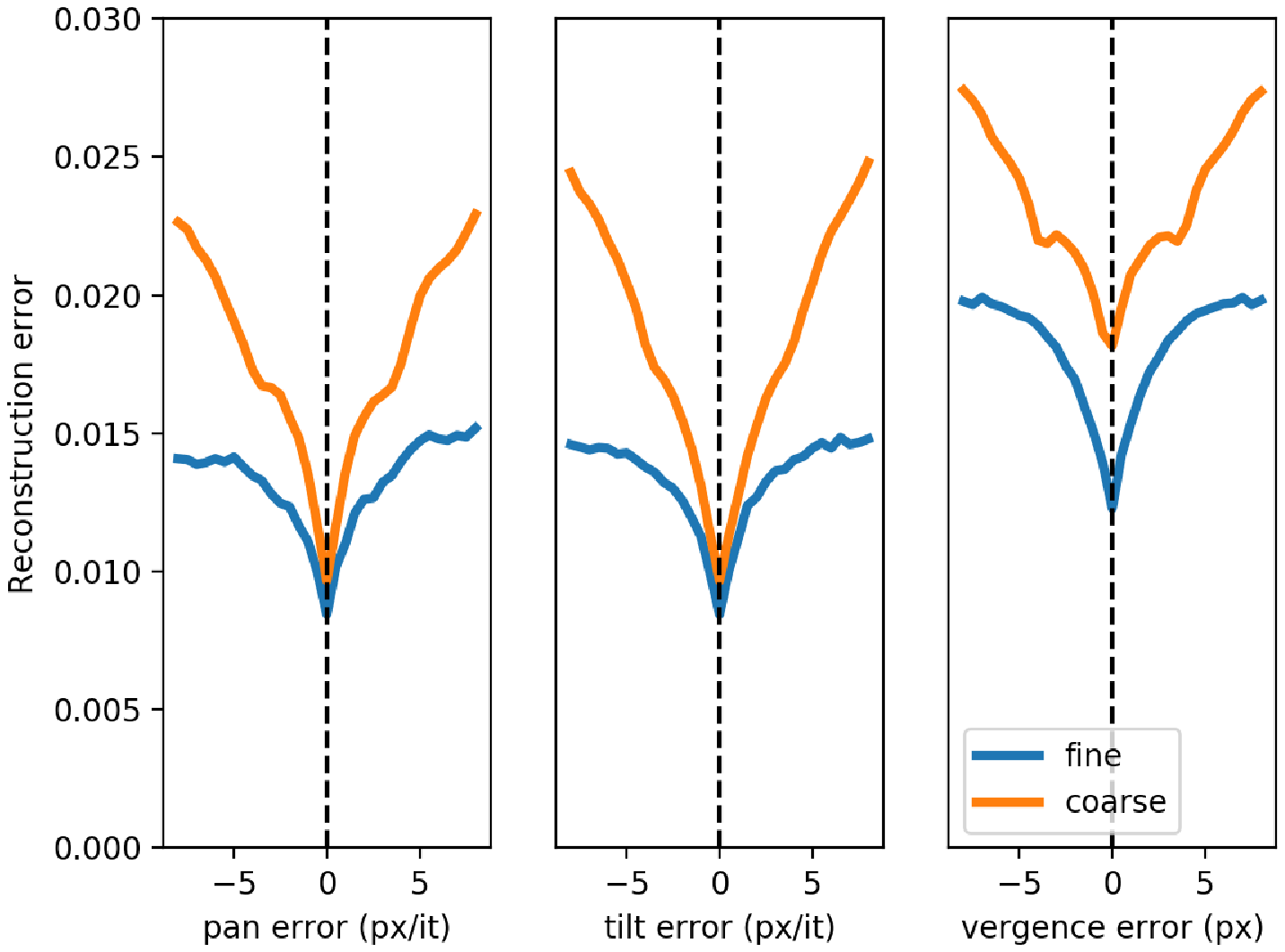}}
\caption{The autoencoders' mean reconstruction errors plotted separately for the two scales as a function of the pan, tilt, and vergence error. For the generation of each plot, the error for the two other joints has been set to $0$.}
\label{vshaped2}
\end{figure}

For successful learning, the pan, tilt and vergence errors must become reflected in the reconstruction errors of the encoders, as explained in Section~\ref{v_shape_text}. We start by analyzing the reconstruction errors of the encoders for every pan, tilt, or vergence error at the end of training with reward function $r^{\rm new}$. Figures~\ref{vshaped1} and \ref{vshaped2} show that for each joint, the reconstruction error is minimal when the absolute joint error is minimal.
In particular, \figurename~\ref{vshaped1} shows the mean reconstruction error for each stimulus displayed on the screen (in blue) and the average for all stimuli (in red), while \figurename~\ref{vshaped2} shows the mean reconstruction error for each scale separately. Repeating the same analysis using random weights for the encoder and decoder shows no difference in the reconstruction quality for low or high absolute errors (the mean error curve is flat instead of being V-shaped, with values around $0.27$, not shown). The characteristic V-shaped curves are a consequence of both learning a compact code of the visual input and adapting the behavior to shape the statistics of the visual input \cite{zhao2012}.

\begin{figure}
\centerline{\includegraphics[width=0.95 \linewidth]{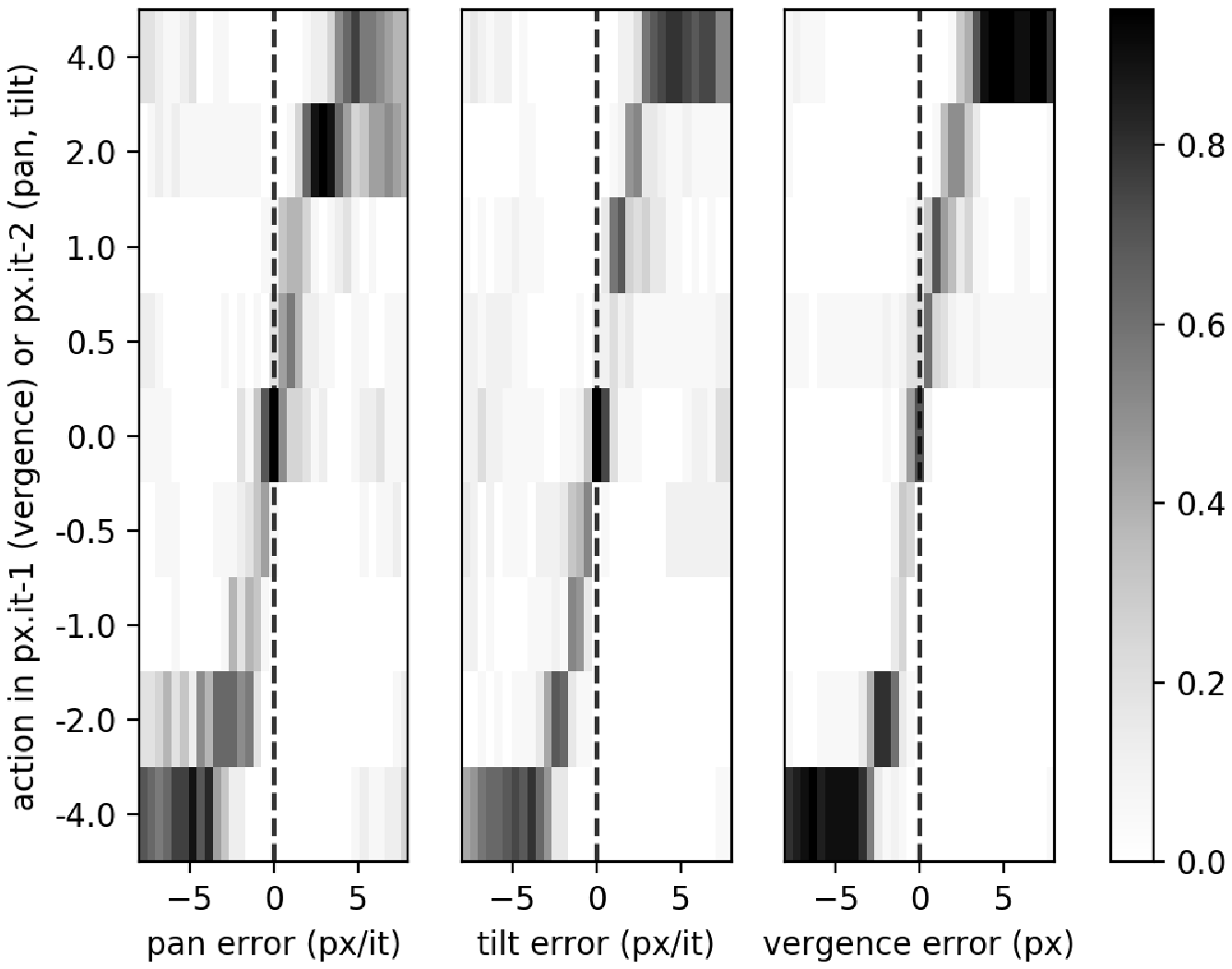}}
\caption{Probability of choosing each action in the action sets as a function of the pan, tilt, and vergence error. For the generation of each plot, the error for the two other joints has been set to $0$.}
\label{policy}
\end{figure}

To show the precision of the learnt policies, we represent for each joint the probability of selecting each possible action in the action set as a function of that joint's error in \figurename~\ref{policy}. The diagonal shapes in the three policies indicate that the model has learned to accurately compensate for any vergence, pan, or tilt errors.

\begin{figure}
\centerline{\includegraphics[width=0.95 \linewidth]{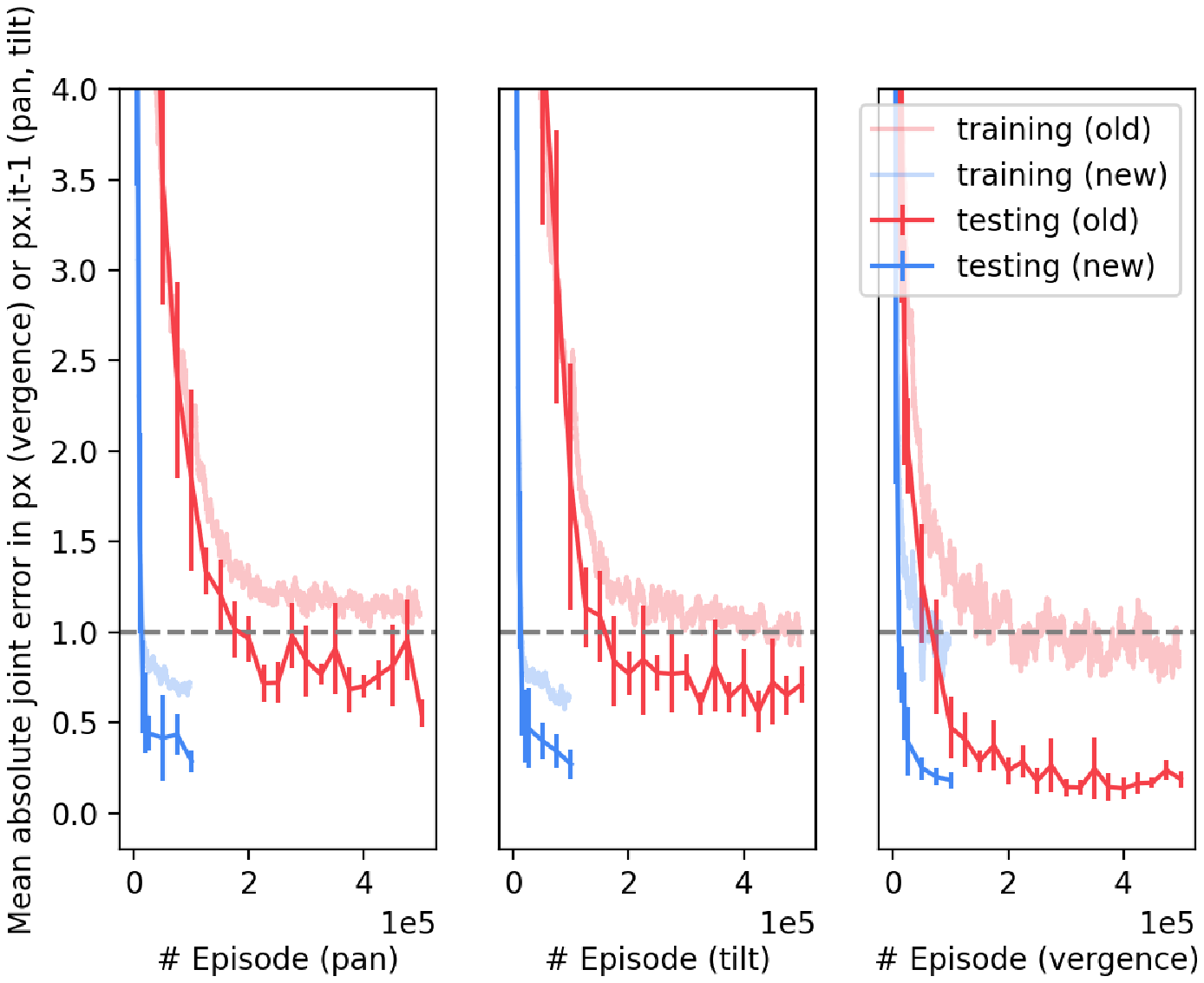}}
\caption{Reduction of errors with training time for the $2$ different rewards. The blue curves correspond to the ``improvement'' reward $ r^{\rm new} $ (Eq.~\ref{new-reward}) and the red curves to $ r^{\rm old} $ (Eq.~\ref{old-reward}). The light curves show the pan, tilt, and vergence errors at the last iteration of an episode as a function of training time when following the $\epsilon$-greedy policies. The data is averaged over 5 independent runs and smoothed for clarity. The dark curves indicate the performance of the greedy policy after \episodelength{} iterations, starting from initial (speed) errors in $\left[-4, 4\right]$ \pxangle{} (vergence) or \pxangularspeedunit{} (pan and tilt), with a step size of half a pixel. Vertical bars indicate the standard deviation across $5$ runs. The testing performance of the new reward is consistently below $1$ \pxangle{} (vergence) or \pxangularspeedunit{} (pan and tilt) after $10,000$ episodes of training demonstrating sub-pixel accuracy.}
\label{joint_errors}
\end{figure}

To show the speed at which the algorithm converges and compare the two reward functions $r^{\rm new}$ and $r^{\rm old}$, \figurename~\ref{joint_errors} plots the average training error at the end of episodes (i.e. the joints' absolute errors while following the $\epsilon$-greedy policies) as a function of training time for both rewards. The testing error is measured at regular intervals as the mean absolute joint error after \episodelength{} iterations of following the greedy policy, starting from initial errors of $-4$, $-2$, $2$ and $4$ \pxangle{} (vergence) or \pxangularspeedunit{} (pan and tilt). From the plots it is evident that the reward $ r^{\rm new} $ measuring the improvement in encoding quality (Eq.~\ref{new-reward}), leads to faster convergence. Focusing on the {\em improvement} of the encoding quality helps the system to deal with very different levels of absolute difficulty for encoding different stimuli (cf.\ Fig.~\ref{vshaped1}).

\begin{figure}
\centerline{\includegraphics[width=0.95 \linewidth]{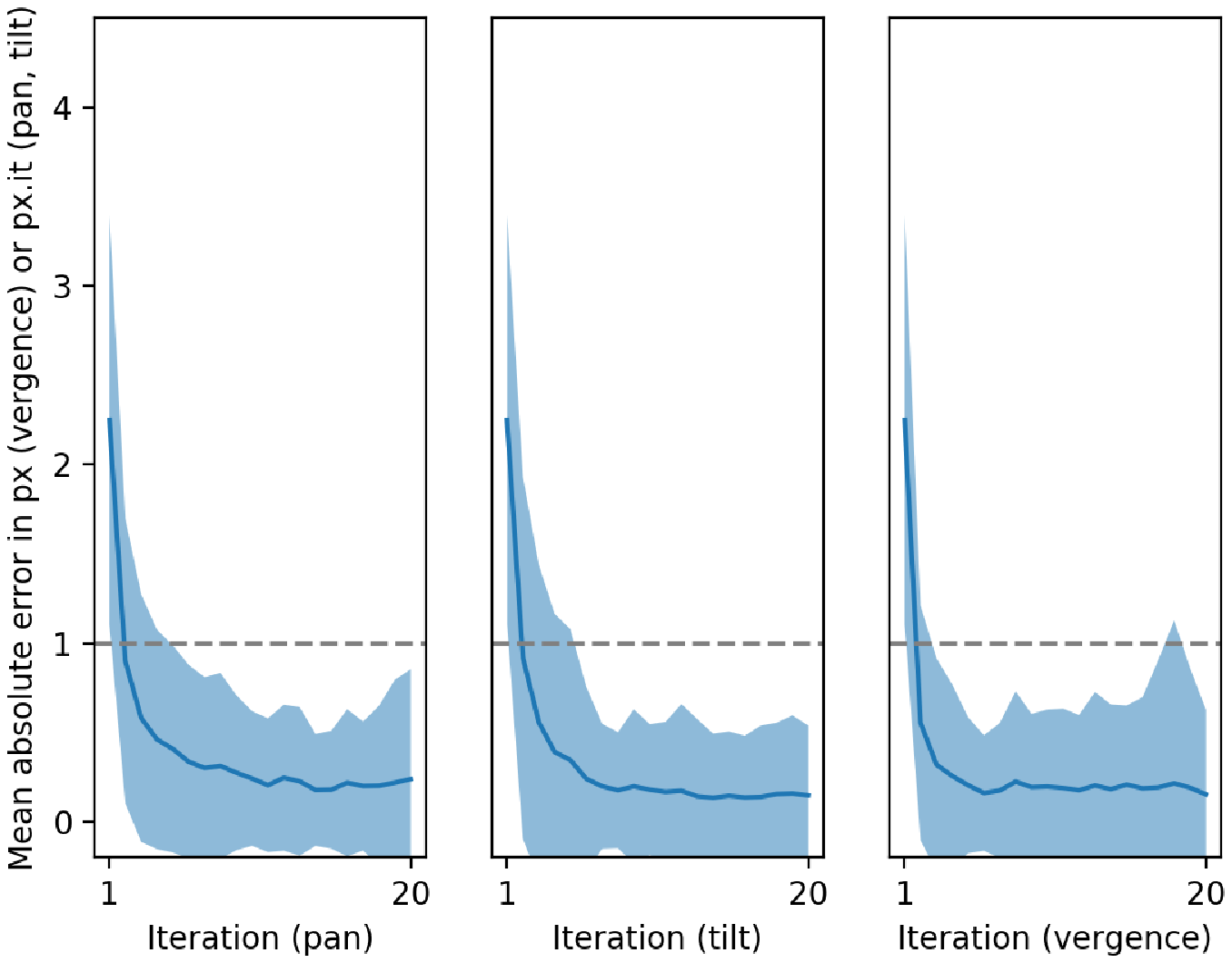}}
\caption{Rapid object fixation and tracking. Pan speed error, tilt speed error, and vergence error are decreasing quickly during one episode and typically reach subpixel levels in one or two steps. The shaded region indicates one standard deviation.}
\label{trajectory}
\end{figure}

Finally, to show how quickly and accurately the algorithm fixates objects and tracks them, \figurename~\ref{trajectory} shows the mean accuracy and its standard deviation, during 20 consecutive iterations, starting from all errors in $\left[-4, 4\right]$ \pxangle{} (vergence) or \pxangularspeedunit{} (pan and tilt) with a step size of half a pixel. Subpixel error levels are typically reached in just one or two iterations despite the discrete action set. Note that an error of, e.g., 3 pixels cannot be reduced to zero in a single step because the two closest discrete actions of 2 pixels and 4 pixels would both lead to a remaining error of 1 pixel.

\section{Discussion}

Understanding the development of the human mind and replicating this development in artificial cognitive agents is a grand challenge for 21$^{\rm st}$ century science. Here, we focused on a very early step in this development, which lays the foundation for most of what follows: the development of early visual representations and the ability to self-calibrate accurate eye movements. From learning to manipulate objects to interacting with social partners, vision is a key sensory modality. In this work, we focused on active stereo and motion vision, proposing a model for their completely autonomous self-calibration. Our work falls inside but also extends the recently proposed Active Efficient Coding (AEC) framework, which is itself an extension of Barlow's classic efficient coding hypothesis\cite{barlow1961} to active perception and therefore rooted in Shannon's information theory. AEC postulates that visual representations and eye movements are jointly optimized to maximize the efficiency of the visual system to encode information. Along these lines,
previous models have shown how AEC can explain the development of active stereo vision \cite{zhao2012,lonini2013}, active motion vision\cite{zhu2017,teuliere2014}, as well as the control of torsional eye movements \cite{zhu2018} and accommodation \cite{eckmann2019} and various combinations thereof, e.g., \cite{narayan2015, lelais2019}. The two key innovations of the present work are to consider ``deeper'' sensory representations compared to the shallow sparse coding approaches used earlier and to use a new intrinsic reward formulation. Regarding the first innovation, we have employed deep convolutional autoencoders for the sensory encoding stage and a Deep Q-Network (DQN) \cite{mnih2013, mnih2016} to map the learned representations onto behavior and obtained very good results. The model quickly achieves sub-pixel accuracy in all degrees of freedom. Regarding the second innovation, we have shown that an intrinsic reward for improvements in encoding quality leads to faster convergence.

From the perspective of biological plausibility, this success comes at a price, however. The learning algorithms used to train the auto-encoder and the DQN rely on error back-propagation mechanisms, which are thought to be not biologically plausible by most researchers in the field. While this should be considered a weakness when evaluating this work as a model of biological mechanisms, it is not problematic from a robotics application perspective.

Another limitation of the approach is that accurate performance requires that the object to be fixated and tracked must be sufficiently big. If the object fills only a fraction of the two regions defining the two spatial scales, then multiple disparities and velocities are present within these regions, because the rest is filled by background. In this case, the system will be ``confused'' and may decide to fixate and stabilize the background. To deal with this problem, a mechanism for foreground/background separation needs to be introduced.

Our work follows the traditional structure of AEC models using a separation into two distinct learning modules --- the first being the deep auto-encoders for unsupervised learning of a sensory representation and generation of reward signals and the second being the DQN for learning behavior via reinforcement learning. It should be questioned however, if this ``hard'' separation is strictly necessary. An alternative architecture might try to blend these functions into a single more homogeneous network. This topic is left for future work.

Another interesting direction for future work is to consider other sensory modalities. Ongoing work (unpublished) is revealing that AEC can also be used to model the self-calibration of echolocation in bats in the auditory modality. More generally, the combination of different sensory modalities is an interesting topic for future research. Arguably, a key step in cognitive development is discovering and understanding the relationships between sensory signals from different modalities in order to arrive at a unified representation of the world. 

\section*{Funding}
This project has received funding from the European Union's Horizon 2020 research and innovation program under grant 713010. J.Triesch is supported by the Johanna Quandt Foundation.

\bibliographystyle{./bibliography/IEEEtran}
\bibliography{./bibliography/IEEEbib.bib}

\begin{thebibliography}{10}
\providecommand{\url}[1]{#1}
\csname url@samestyle\endcsname
\providecommand{\newblock}{\relax}
\providecommand{\bibinfo}[2]{#2}
\providecommand{\BIBentrySTDinterwordspacing}{\spaceskip=0pt\relax}
\providecommand{\BIBentryALTinterwordstretchfactor}{4}
\providecommand{\BIBentryALTinterwordspacing}{\spaceskip=\fontdimen2\font plus
\BIBentryALTinterwordstretchfactor\fontdimen3\font minus
  \fontdimen4\font\relax}
\providecommand{\BIBforeignlanguage}[2]{{%
\expandafter\ifx\csname l@#1\endcsname\relax
\typeout{** WARNING: IEEEtran.bst: No hyphenation pattern has been}%
\typeout{** loaded for the language `#1'. Using the pattern for}%
\typeout{** the default language instead.}%
\else
\language=\csname l@#1\endcsname
\fi
#2}}
\providecommand{\BIBdecl}{\relax}
\BIBdecl

\bibitem{zhao2013}
\BIBentryALTinterwordspacing
L.~Lonini, S.~Forestier, C.~Teulière, Y.~Zhao, B.~Shi, and J.~Triesch,
  ``Robust active binocular vision through intrinsically motivated learning,''
  \emph{Frontiers in Neurorobotics}, vol.~7, p.~20, 2013. [Online]. Available:
  \url{https://www.frontiersin.org/article/10.3389/fnbot.2013.00020}
\BIBentrySTDinterwordspacing

\bibitem{bourdonnaye2018}
F.~D.~L. Bourdonnaye, C.~Teuli{\`e}re, J.~Triesch, and T.~Chateau, ``Stage-wise
  learning of reaching using little prior knowledge,'' \emph{Front. Robotics
  and AI}, vol. 2018, 2018.

\bibitem{jeffries}
P.~J. Jeffries~AM, Killian~NJ, ``Mapping the primate lateral geniculate
  nucleus: a review of experiments and methods.'' \emph{J Physiol Paris}, 2014.

\bibitem{pavro}
X.~Xu, J.~M. Ichida, J.~D. Allison, J.~D. Boyd, A.~Bonds, and V.~A. Casagrande,
  ``A comparison of koniocellular, magnocellular and parvocellular receptive
  field properties in the lateral geniculate nucleus of the owl monkey (aotus
  trivirgatus),'' \emph{The Journal of physiology}, vol. 531, no.~1, pp.
  203--218, 2001.

\bibitem{binocular}
N.~Qian, ``Binocular disparity and the perception of depth,'' \emph{Neuron},
  vol.~18, no.~3, pp. 359--368, 1997.

\bibitem{motion}
A.~Borst and M.~Egelhaaf, ``Principles of visual motion detection,''
  \emph{Trends in neurosciences}, vol.~12, no.~8, pp. 297--306, 1989.

\bibitem{simplecells1}
J.~Y. Lettvin, H.~R. Maturana, W.~S. McCulloch, and W.~H. Pitts, ``What the
  frog's eye tells the frog's brain,'' \emph{Proceedings of the IRE}, vol.~47,
  no.~11, pp. 1940--1951, 1959.

\bibitem{simplecells2}
B.~Scholl, J.~Burge, and N.~J. Priebe, ``Binocular integration and disparity
  selectivity in mouse primary visual cortex,'' \emph{Journal of
  neurophysiology}, vol. 109, no.~12, pp. 3013--3024, 2013.

\bibitem{barlow1961}
H.~Barlow, ``Possible principles underlying the transformations of sensory
  messages,'' \emph{Sensory Communication}, vol.~1, 01 1961.

\bibitem{zhaoping2007}
L.~Zhaoping, ``Theoretical understanding of the early visual processes by data
  compression and data selection,'' \emph{Network (Bristol, England)}, vol.~17,
  pp. 301--34, 01 2007.

\bibitem{kelly1962}
D.~H. Kelly, ``Information capacity of a single retinal channel,'' \emph{IRE
  Trans. Information Theory}, vol.~8, pp. 221--226, 1962.

\bibitem{nirenberg2001}
\BIBentryALTinterwordspacing
S.~Nirenberg, S.~M. Carcieri, A.~L. Jacobs, and P.~E. Latham, ``Retinal
  ganglion cells act largely as independent encoders,'' \emph{Nature}, vol.
  411, no. 6838, pp. 698--701, 2001. [Online]. Available:
  \url{https://doi.org/10.1038/35079612}
\BIBentrySTDinterwordspacing

\bibitem{teuliere2014}
T.~N. {Vikram}, C.~{Teulière}, C.~{Zhang}, B.~E. {Shi}, and J.~{Triesch},
  ``Autonomous learning of smooth pursuit and vergence through active efficient
  coding,'' in \emph{4th International Conference on Development and Learning
  and on Epigenetic Robotics}, 2014, pp. 448--453.

\bibitem{eckmann2019}
\BIBentryALTinterwordspacing
S.~Eckmann, L.~Klimmasch, B.~E. Shi, and J.~Triesch, ``Active efficient coding
  explains the development of binocular vision and its failure in amblyopia,''
  \emph{Proceedings of the National Academy of Sciences}, vol. 117, no.~11, pp.
  6156--6162, 2020. [Online]. Available:
  \url{https://www.pnas.org/content/117/11/6156}
\BIBentrySTDinterwordspacing

\bibitem{mnih2013}
V.~Mnih, K.~Kavukcuoglu, D.~Silver, A.~Graves, I.~Antonoglou, D.~Wierstra, and
  M.~Riedmiller, ``Playing atari with deep reinforcement learning,''
  \emph{arXiv preprint arXiv:1312.5602}, 2013.

\bibitem{mnih2016}
V.~Mnih, A.~P. Badia, M.~Mirza, A.~Graves, T.~Lillicrap, T.~Harley, D.~Silver,
  and K.~Kavukcuoglu, ``Asynchronous methods for deep reinforcement learning,''
  in \emph{International conference on machine learning}, 2016, pp. 1928--1937.

\bibitem{huber1964}
\BIBentryALTinterwordspacing
P.~J. Huber, ``Robust estimation of a location parameter,'' \emph{Ann. Math.
  Statist.}, vol.~35, no.~1, pp. 73--101, 03 1964. [Online]. Available:
  \url{https://doi.org/10.1214/aoms/1177703732}
\BIBentrySTDinterwordspacing

\bibitem{james2019}
S.~James, M.~Freese, and A.~J. Davison, ``Pyrep: Bringing v-rep to deep robot
  learning,'' \emph{arXiv preprint arXiv:1906.11176}, 2019.

\bibitem{olmos2004}
\BIBentryALTinterwordspacing
A.~Olmos and F.~A.~A. Kingdom, ``A biologically inspired algorithm for the
  recovery of shading and reflectance images,'' \emph{Perception}, vol.~33,
  no.~12, pp. 1463--1473, 2004, pMID: 15729913. [Online]. Available:
  \url{https://doi.org/10.1068/p5321}
\BIBentrySTDinterwordspacing

\bibitem{adam}
D.~P. Kingma and J.~Ba, ``Adam: A method for stochastic optimization,''
  \emph{arXiv preprint arXiv:1412.6980}, 2014.

\bibitem{zhao2012}
Y.~Zhao, C.~A. Rothkopf, J.~Triesch, and B.~E. Shi, ``A unified model of the
  joint development of disparity selectivity and vergence control,'' in
  \emph{2012 IEEE International Conference on Development and Learning and
  Epigenetic Robotics (ICDL)}.\hskip 1em plus 0.5em minus 0.4em\relax IEEE,
  2012, pp. 1--6.

\bibitem{lonini2013}
L.~{Lonini}, Y.~{Zhao}, P.~{Chandrashekhariah}, B.~E. {Shi}, and J.~{Triesch},
  ``Autonomous learning of active multi-scale binocular vision,'' in \emph{2013
  IEEE Third Joint International Conference on Development and Learning and
  Epigenetic Robotics (ICDL)}, 2013, pp. 1--6.

\bibitem{zhu2017}
Q.~Zhu, J.~Triesch, and B.~E. Shi, ``Autonomous, self-calibrating binocular
  vision based on learned attention and active efficient coding,'' \emph{2017
  Joint IEEE International Conference on Development and Learning and
  Epigenetic Robotics (ICDL-EpiRob)}, pp. 27--32, 2017.

\bibitem{zhu2018}
\BIBentryALTinterwordspacing
Q.~Zhu, C.~Zhang, J.~Triesch, and B.~E. Shi, ``Autonomous learning of
  cyclovergence control based on active efficient coding,'' in \emph{2018 Joint
  {IEEE} 8th International Conference on Development and Learning and
  Epigenetic Robotics, ICDL-EpiRob 2018, Tokyo, Japan, September 17-20,
  2018}.\hskip 1em plus 0.5em minus 0.4em\relax {IEEE}, 2018, pp. 251--256.
  [Online]. Available: \url{https://doi.org/10.1109/DEVLRN.2018.8761033}
\BIBentrySTDinterwordspacing

\bibitem{narayan2015}
A.~{Priamikov}, V.~{Narayan}, B.~E. {Shi}, and J.~{Triesch}, ``The role of
  contrast sensitivity in the development of binocular vision: A computational
  study,'' in \emph{2015 Joint IEEE International Conference on Development and
  Learning and Epigenetic Robotics (ICDL-EpiRob)}, 2015, pp. 33--38.

\bibitem{lelais2019}
A.~Lelais, J.~Mahn, V.~Narayan, C.~Zhang, B.~E. Shi, and J.~Triesch,
  ``Autonomous development of active binocular and motion vision through active
  efficient coding,'' \emph{Frontiers in neurorobotics}, vol.~13, p.~49, 2019.

\end{thebibliography}

\end{document}